\documentclass[a4paper,fleqn]{cas-dc}

\usepackage[authoryear,longnamesfirst]{natbib}

\def\tsc#1{\csdef{#1}{\textsc{\lowercase{#1}}\xspace}}
\tsc{WGM}
\tsc{QE}
\tsc{EP}
\tsc{PMS}
\tsc{BEC}
\tsc{DE}

\usepackage{pifont}
\usepackage{amsmath}
\usepackage{color}
\usepackage{caption}
\usepackage{hyperref}

\begin{document}
\let\WriteBookmarks\relax
\def\floatpagepagefraction{1}
\def\textpagefraction{.001}
\let\printorcid\relax
\shortauthors{Chun Bao et~al.}

\title [mode = title]{Improved Dense Nested Attention Network Based on Transformer for Infrared Small Target Detection}

\cortext[1]{Corresponding author.}

\credit{Conceptualization of this study, Methodology, Software}

\author[1]{Chun Bao}

\author[1]{Jie Cao}
\cormark[1]

\author[1]{Yaqian Ning}
\author[1]{Tianhua Zhao}
\author[1]{Zhijun Li}
\author[1]{Zechen Wang}
\author[1]{Li Zhang}
\author[1]{Qun Hao}

\credit{Data curation, Writing - Original draft preparation}

\address[1]{Key Laboratory of Biomimetic Robots and Systems, Ministry of Education, Beijing Institute of Technology, Beijing 100081, China}

\begin{abstract}
Infrared small target detection based on deep learning offers unique advantages in separating small targets from complex and dynamic backgrounds. However, the features of infrared small targets gradually weaken as the depth of convolutional neural network (CNN) increases. To address this issue, we propose a novel method for detecting infrared small targets called improved dense nested attention network (IDNANet), which is based on the transformer architecture. We preserve the dense nested structure of dense nested attention network (DNANet) and introduce the Swin-transformer during feature extraction stage to enhance the continuity of features. Furthermore, we integrate the ACmix attention structure into the dense nested structure to enhance the features of intermediate layers. Additionally, we design a weighted dice binary cross-entropy (WD-BCE) loss function to mitigate the negative impact of foreground-background imbalance in the samples. Moreover, we develop a dataset specifically for infrared small targets, called BIT-SIRST. The dataset comprises a significant amount of real-world targets and manually annotated labels, as well as synthetic data and corresponding labels. We have evaluated the effectiveness of our method through experiments conducted on public datasets. In comparison to other state-of-the-art methods, our approach outperforms in terms of probability of detection ($P_d$), false-alarm rate ($F_a$), and mean intersection of union ($mIoU$). The $mIoU$ reaches 90.89\% on the NUDT-SIRST dataset and 79.72\% on the SIRST dataset. The BIT-SIRST dataset and codes are available openly at \href{https://github.com/EdwardBao1006/bit\_sirst}{\color[HTML]{B22222}{https://github.com/EdwardBao1006/bit\_sirst}}.
\end{abstract}


\maketitle

\begin{figure*}
	\centering
	\includegraphics[width=0.95\textwidth]{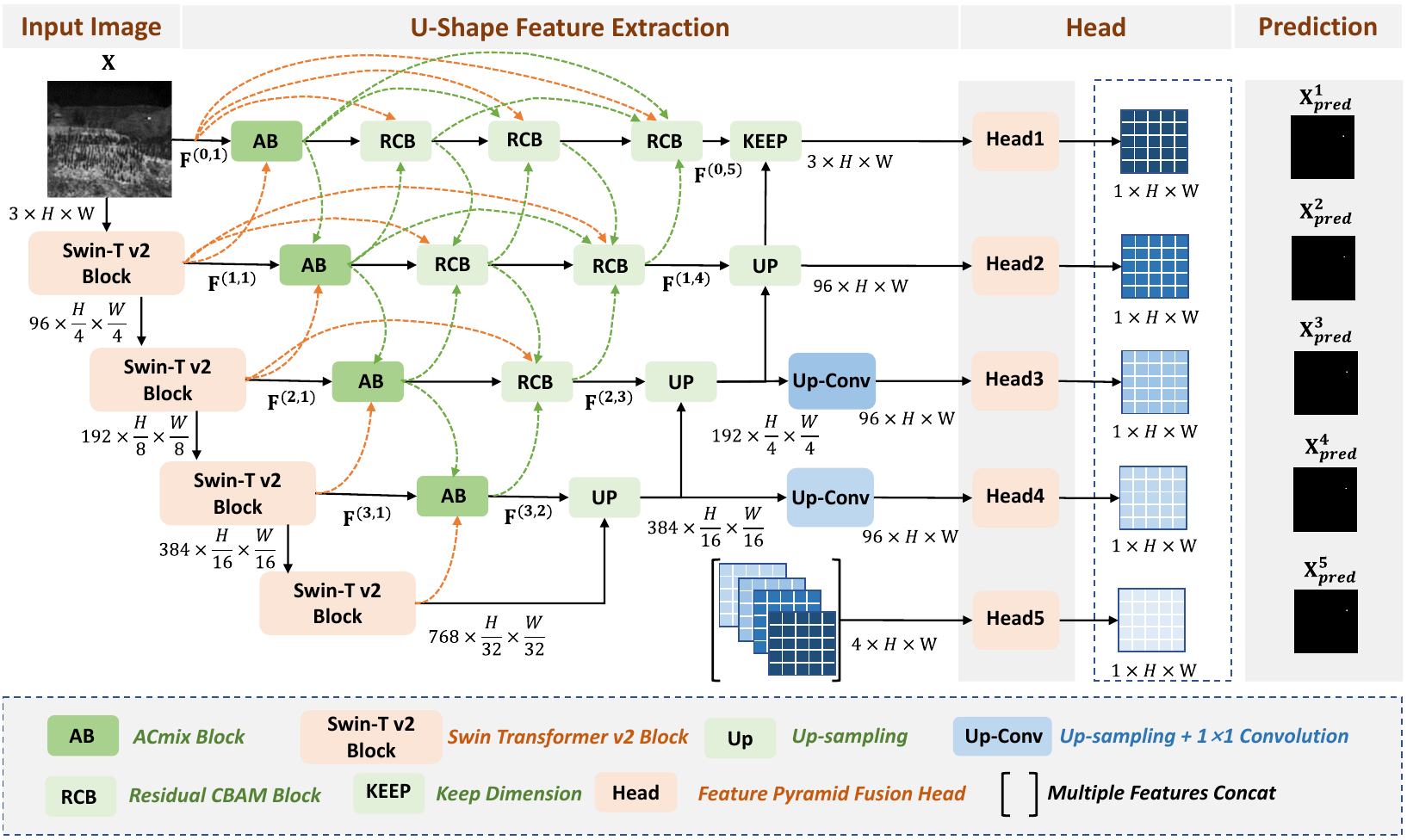}
	\caption{The overview architecture of the IDNANet. The overall network mainly consists of several components: \textit{Input Image}, \textit{U-Shape Feature Extraction}, \textit{Head}, and \textit{Prediction}. Where the input infrared image ${\bf{X}} \in {{\mathbb{R}}^{3\times H \times W}}$. ${{\bf{F}}^{(i,j)}}$ represents the feature map of the U-Shape network at position $(i, j)$. $H$ and $W$ represent the height and width of the image, respectively.}
	\label{fig1}
\end{figure*}

\section{Introduction}
In the past decade, the utilization of infrared small target detection has been extensively applied in diverse domains, encompassing military and space remote sensing \citep{wang2020ship, nian2023local, wu2022srcanet}. From the analysis of infrared small targets, two primary attributes are observable. \textbf{Dim}: infrared small targets have a low signal-to-noise ratio and are easily overwhelmed by complex and variable backgrounds. \textbf{Small}: the target size in infrared images is relatively small. Furthermore, infrared small targets lack feature information such as color and texture. These two distinctive characteristics render the detection of infrared small targets notably arduous \citep{kou2023infrared}. Specifically, this includes: \textbf{(1) The features available for infrared small targets are constrained.} The small size of the target in the image and the varied distribution in the natural scene make it challenging to employ a unified mathematical model to depict the target \citep{ying2023mapping}. \textbf{(2) The signal-to-noise ratio of infrared images is feeble.} Due to the extensive imaging distance, the small targets are susceptible to being overshadowed and disrupted by the clutter and noise characteristics of clouds and ocean waves in the background\citep{dai2023one}. \textbf{(3) The backgrounds of images containing small infrared targets are typically intricate and raucous.} Consequently, the detection of infrared small targets has emerged as a prominent research topic and various methods have been proposed. These methods can be divided into traditional and deep learning based (DL-based) methods. Traditional methods include filter-based, human visual system-based (HVS-based), and image data structure-based (IDS-based) approaches \citep{xu2023infrared, zhang2023infrared, lu2023infrared}. Deep learning methods can be categorized into detection-based and segmentation based methods \citep{du2021spatial, pan2023abc, chen2022local}. The earliest traditional approach is the filtering-based algorithm, which detects small targets by comparing image gray information and visual saliency. This method estimates the background of the infrared image and suppresses background noise using filters like Top-Hat \citep{rivest1996detection},  Max-Median \citep{deshpande1999max}, and Median Filtering. The human visual system-based method relies on changes in local texture caused by small targets to locate their position and morphology, such as local contrast measure (LCM) \citep{chen2013local}. However, this method is time-consuming, prone to block effects, and not suitable for detecting dark targets. The image structure-based method transforms small target detection into a convex optimization problem, considering the sparsity of the object and the low-rank nature of the background \citep{zhang2019infrared}. For example, the infrared patch-image (IPI) \citep{gao2013infrared} generalizes the traditional infrared image model to a new infrared block image model using a local block construction method. Currently, single-stage \citep{liu2016ssd,wang2023yolov7} and two-stage \citep{girshick2015fast,ren2015faster} deep learning algorithms are widely used for object detection, but they have limitations in detecting small targets, especially infrared small targets. Specialized infrared small target detectors like asymmetric contextual modulation (ACM)  \citep{dai2021asymmetric} and attention-guided pyramid context network (AGPCNet) \citep{zhang2021agpcnet} have been developed to address this issue. Although deep learning methods have improved, the increase in CNN layers has weakened the ability to detect small infrared targets, resulting in reduced accuracy. Thus, the dense nested attention network (DNANet) \citep{li2022dense} introduces a dense nested infrared small target detection network to address these problems and enhance the utilization of multi-level features. Inspired by DNANet, we develop an improved dense nested attention network (IDNANet), which is based on the dense nested structure. Specifically, we enhance the feature extraction network of DNANet by replacing the ResNet with the Swin-transformer (Swin-T) v2 \citep{liu2022swin} network. Compared to convolutional neural networks (CNNs), Swin-T can capture global information by utilizing the self-attention mechanism to establish long-distance dependencies. Furthermore, we introduce the ACmix \citep{pan2022integration} attention mechanism and design the ACmix block ($\bf{AB}$) as an inter-layer feature enhancement module. To optimize the loss, we propose the weighted dice binary cross-entropy (WD-BCE) loss function. WD-BCE assigns weights to the calculated loss values, mitigating the negative impact of foreground-background imbalance and improving the stability of network training \citep{li2019dice,bruch2019analysis}. Additionally, we develop a novel single-frame infrared small target (SIRST) detection dataset called BIT-SIRST. In summary, the contributions of this paper can be summarized as follows.

\begin{itemize}
	\item \textbf{We enhance the DNANet and propose a novel transformer based method for infrared small target detection called IDNANet.} While preserving the dense nested structure of DNANet, we incorporate Swin-T v2 to improve feature continuity. Additionally, we integrate the ACmix attention structure into the dense nested structure to ensure that the features extracted by the transformer are better suited for the dense nested structure. 
	\item \textbf{We formulate a novel loss function by recombining Dice and BCE loss functions, which we refer to as WD-BCE.} This loss function addresses the negative impact caused by foreground-background imbalance in the samples and ensures equal calculation of the loss for each pixel. It promotes network convergence and improves overall performance.
	\item \textbf{We create a unique single-frame dataset for infrared small target detection called BIT-SIRST.} This dataset combines both synthetic and real data. A portion of the dataset includes complete truth data with retained contour information. 
\end{itemize}

\section{Related Works}
\subsection{Single Frame Infrared Small Target Detection}
Traditional methods for infrared small target detection primarily rely on filter-based approaches to suppress back-ground noise in images \citep{moradi2018false, han2019local}. Nonetheless, traditional methods encounter various challenges in practice. For instance, the manual design hyperparameters need synchronous modification when switching between different scenarios \citep{han2014robust, liu2018tiny}. Unlike traditional image-based detection methods, deep learning approaches based on data-driven techniques can effectively address these challenges \citep{han2022kcpnet, wang2022interior}. MDvsFA \citep{wang2019miss} introduces a conditional generative adversarial network framework, decomposing the infrared small target detection task into two subtasks: MD minimization and FA minimization. ALCNet\citep{dai2021attentional} combines discriminative networks with conventional model-driven methods, utilizing data and image region knowledge. By modularizing the traditional local contrast measurement method into a deep parameter-free nonlinear feature refinement layer, ALCNet incorporates a feature map cyclic shift scheme to enable context interactions over longer distances with clear physical interpretability. DNANet \citep{li2022dense} designs a dense nested structure for the U-shaped network, enhancing information interaction between low-level and high-level networks while preserving the original characteristics of infrared small targets and improving communication and fusion between features of different scales. UIU-Net \citep{wu2022uiu} extends the capabilities of the U-Net backbone, enabling multi-level and multi-scale representation learning of targets. In recent studies, researchers have also utilized Transformers in infrared small target detection. For example, MTU-Net \citep{wu2023mtu} incorporates ViT \citep{dosovitskiy2020image} into U-Net to extract multi-level features, exploiting the long-distance dependencies of small targets through coarse-to-fine feature extraction and multi-level feature fusion. In contrast to the aforementioned works, our approach combines the strengths of the U-shaped backbone network and the dense nested structure of DNANet to facilitate multi-scale feature interaction across different information layers. We integrate the Swin-T structure for more efficient extraction of infrared small target features. By designing the WD-BCE loss function, we can significantly enhance the performance of infrared small target detection.

\subsection{Dataset for Infrared Small Target Detection}
The data-driven deep learning network has significantly improved the detection performance of infrared small targets. However, the model requires a large-scale training dataset to meet high requirements. The quality and quantity of the training dataset directly determine the detection performance of the model. Although the ROC curve provides a more comprehensive reflection of the algorithm's detection performance, it requires a more extensive dataset to be universally applicable. In recent years, various infrared small target detection datasets for different scenes and targets have been proposed. SIRST aims to extract the most representative images from the sequence. This dataset contains a total of 427 infrared images, which include 480 targets. Many of these targets are blurry and hidden in complex backgrounds. The NUST-SIRST dataset consists of 11 real infrared sequences with 2098 frames. It also contains 100 real independent infrared images with different small targets. CQU-SIRST is a synthetic dataset that includes 1676 images. NUDT-SIRST is another synthetic dataset, consisting of over 1327 images. NUDT-SIRST-Sea is an extension of the previous dataset, specifically designed for ship targets on the sea surface. It comprises 48 real images captured from near-infrared and short-infrared bands by sensors mounted on low Earth orbit satellites. Although public datasets have contributed to the advancement of infrared small target detection, they still present challenges such as a limited field of view, complex backgrounds, weak target features, and varying sizes. Synthetic datasets can help address the challenges and mitigate the problem of manual label misalignment in real datasets. However, achieving greater authenticity is also necessary for synthetic datasets. Therefore, our BIT-SIRST dataset combines real data with manually crafted labels and real backgrounds with synthetic target information. This combination enhances the diversity of scenes and targets in the infrared small target dataset while ensuring the accuracy of the labels.

\section{Improved Dense Nested Attention Network Based on Transformer}
\subsection{Overall Structure}
The structure of the IDNANet is shown in Fig. \ref{fig1}. This structure preserves the dense nested structure of DNANet and utilizes a U-shape CNN for feature extraction. In contrast to DNANet, we have enhanced the original residual CNN with Swin-T. With a conventional CNN, the pooling layer would weaken the target feature in each sample. However, Swin-T employs a feature encoder-decoder approach, effectively preserving the image characteristics of infrared small targets. In Fig. \ref{fig1}, we can observe that we have enhanced the residual CBAM block ($\bf{RCB}$) used in DNANet to the ACmix block. ACmix offers the advantage of combining the traditional CNN-based attention characteristics with the transformer-based characteristics when aggregating different attention information in the model. Since the image features used for feature extraction are derived from the Swin-T network, relying solely on the attention module of CNN would result in unmatched information during feature aggregation across different layers. Following the low-level feature aggregation in the $\bf{AB}$ module, the U-shaped feature extraction network retains the dense nested design of DNANet in its middle structure. In this middle feature extraction module, we also use the $\bf{RCB}$. This retention of structure aims to reduce the computational workload in the middle layer. We then take the feature maps of different scales obtained from the dense nested layers and convert them into high-level feature maps with the same size but differing channel numbers through up-sampling ($\bf{UP}$) and the keep-size operation ($\bf{KEEP}$). In the Head part, we utilize the feature pyramid fusion structure. This structure converts the obtained feature maps into five prediction result maps of ${1\times H \times W}$ size. After the U-shaped feature extraction network and the Head part, we obtain five prediction maps. During the training, we calculate the WD-BCE loss values between these five prediction maps and the corresponding labels.

\subsection{Swin-T Block}
\begin{figure*}
	\centering
	\includegraphics[width=0.75\textwidth]{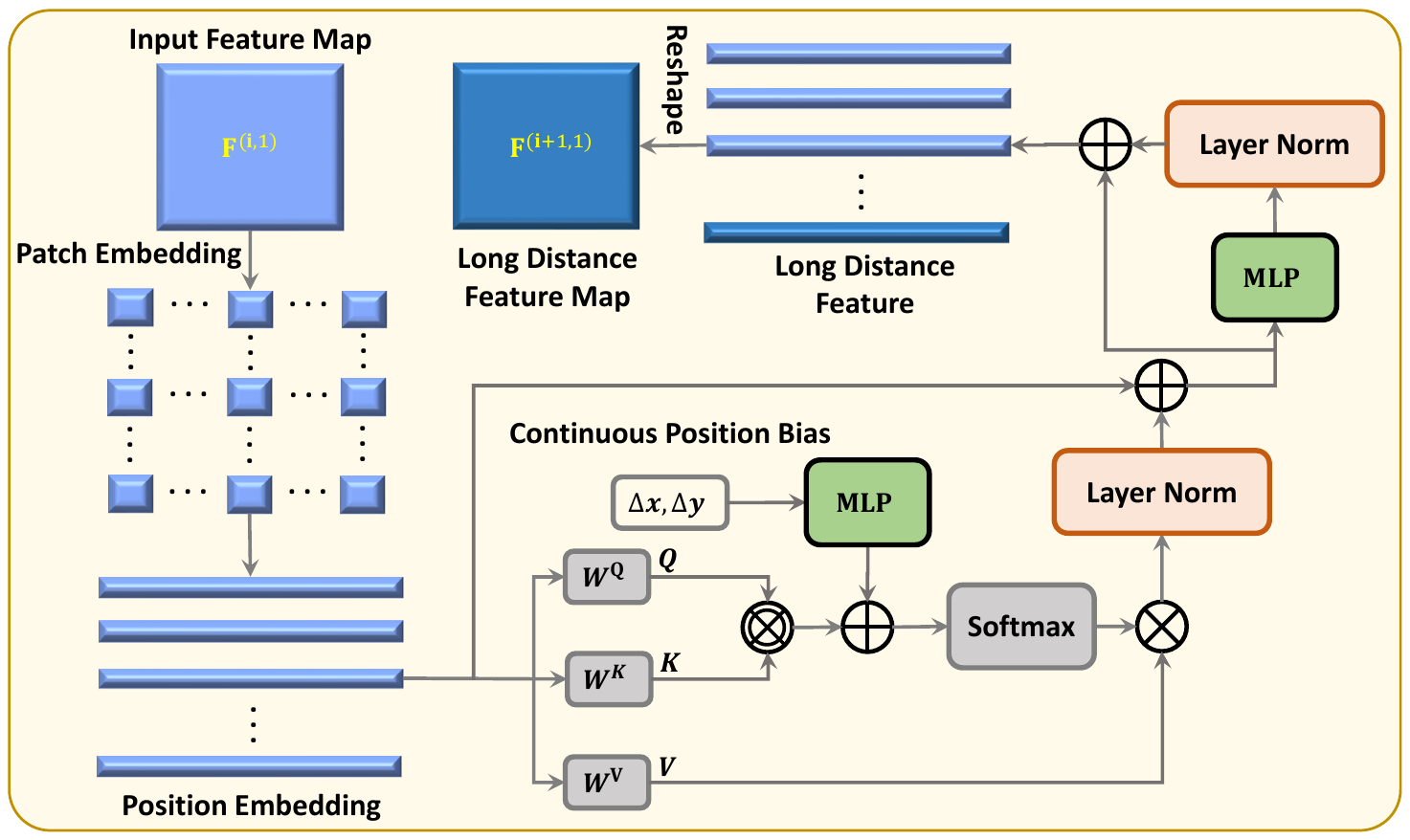}
	\caption{The structure of the Swin-T block. In light of the shortcomings of feature dissipation that arise in the feature extraction stage when using CNN-based methods, we utilize the backbone network of Swin-T v2 in this study. Specifically, we conduct patch embedding and position embedding operations on infrared images containing small targets. The processed image is then fed into the standard transformer process of encoder and decoder.}
	\label{fig2}
\end{figure*}

The Swin-T module used in our feature extraction stage is shown in Fig. \ref{fig2}. Based on the powerful modeling ability of Transformer units, visual signal prior makes it friendly to various visual tasks. Swin-T has two features: hierarchical feature structure and linear complexity. This hierarchical feature structure makes Swin-T very suitable for models with FPN and U-Net structures. In our IDNANet, not only does it contain FPN, but it also improves upon the basis of U-Net. Considering the accuracy improvement brought by Swin-T as a backbone, we ultimately improved the traditional residual-based CNN network into a module structure of Swin-T. In this module, we first divide the input image into non-overlapping patches. Each patch is considered as a token, and its feature is the concatenation of all pixel RGB values. Modified Swin Transformer blocks are applied to these patches. To produce a hierarchical representation, the number of patches is reduced through patch merging layers as the network deepens. In the encoding position, we use Log-spaced coordinates to obtain relative position encoding coordinates ($\Delta x$,$\Delta y$), replacing linear transformation with logarithmic transformation. It is worth noting that in Swin-T V2, the operation of $\textit{Q}$ and $\textit{K}$ has changed from dot product operation to scaled cosine operation. Scaled cosine ensures that the computation is independent of the amplitude of the block input, and the network will not fall into extreme states.

\subsection{ACmix Block}
\begin{figure*}
	\centering
	\includegraphics[width=0.75\textwidth]{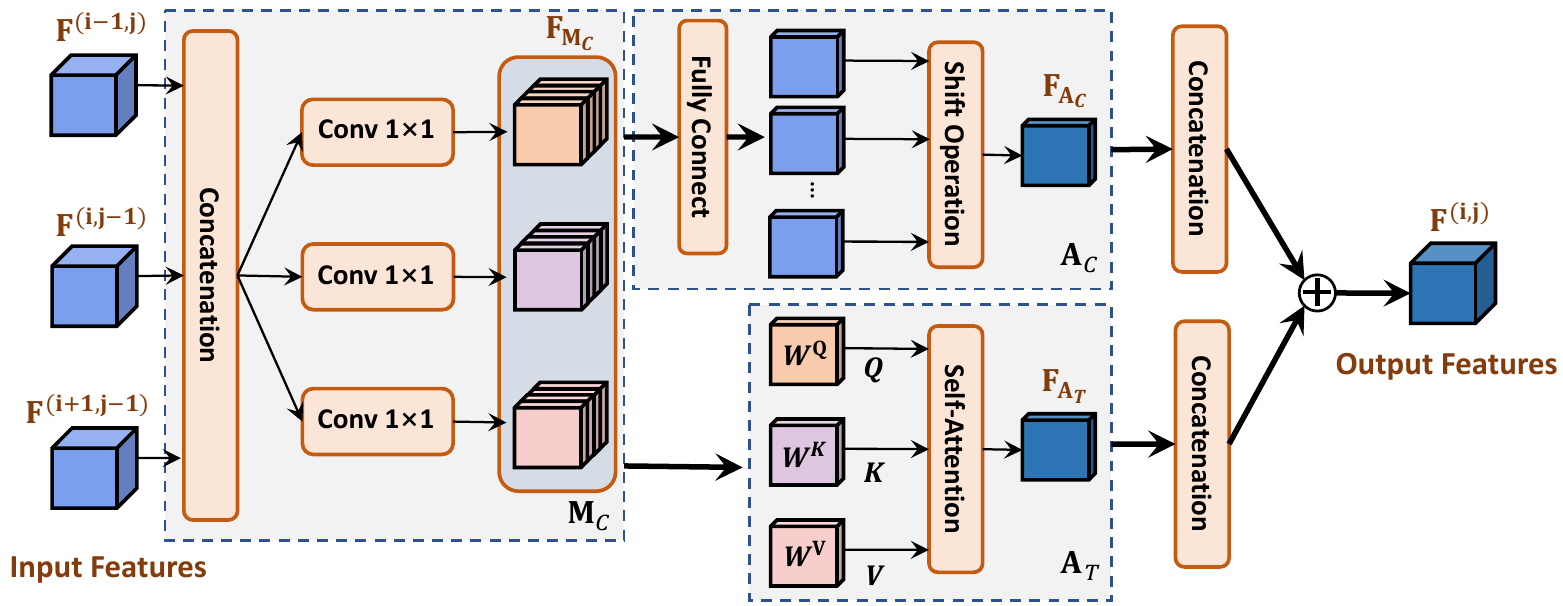}
	\caption{The architecture of ACmix block. In this block, we employ the feature maps of adjacent components of each node as input. This approach not only facilitates cross-layer information exchange between feature maps but also enhances features extracted by Swin-T block.}
	\label{fig3}
\end{figure*}

After the feature extraction process, we generate multi-scale outputs for the feature maps of the U-shape network. To ensure that the network can effectively handle targets of varying sizes during training, we extract the ${{\bf{F}}^{(0,5)}}$, ${{\bf{F}}^{(1,4)}}$, ${{\bf{F}}^{(2,3)}}$ and ${{\bf{F}}^{(3,2)}}$ feature maps separately before performing pyramid fusion on the feature maps. In subsequent experiments, we also validate the effectiveness of extracting these feature maps for network training. Here, ${{\bf{F}}^{(0,1)}}$, ${{\bf{F}}^{(1,1)}}$, ${{\bf{F}}^{(2,1)}}$ and ${{\bf{F}}^{(3,1)}}$ are the output feature maps of $\bf{AB}$. The calculation of $\bf{AB}$ is shown in Fig. \ref{fig3}, we assume that the current input feature map is ${{\bf{F}}^{(i,j-1)}}$, the feature map extracted by the previous layer is ${{\bf{F}}^{(i-1,j)}}$, and the feature map of the next layer is ${{\bf{F}}^{(i+1,j-1)}}$. Here, we concatenate the feature maps of adjacent layers into three ways later for channel separation ${{\bf{M}}_C}$ operation, as shown in Equation \ref{eq1}. After obtaining the channel separation features, we further perform the convolutional attention-based calculation ${{\bf{A}}_C}$ and the transformer-based self-attention calculation ${{\bf{A}}_T}$, as illustrated in Equations \ref{eq2} and \ref{eq3}. And the intermediate feature maps ${\bf{F}}_{{{\bf{A}}_C}}^{(i,j)}$ and ${\bf{F}}_{{{\bf{A}}_T}}^{(i,j)}$ are calculated respectively. Finally, the ACmix multi-scale attention feature map ${{\bf{F}}^{(i,j)}}$ is obtained by adding as shown in Equation \ref{eq4}.

\begin{equation}\label{eq1}
	{\bf{F}}_{{{\bf{M}}_C}}^{(i,j)} = {{\bf{M}}_C}({{\bf{F}}^{(i - 1,j)}},{{\bf{F}}^{(i,j - 1)}},{{\bf{F}}^{(i + 1,j - 1)}})
\end{equation}

\begin{equation}\label{eq2}
	{\bf{F}}_{{{\bf{A}}_C}}^{(i,j)} = {{\bf{A}}_C}({\bf{F}}_{{{\bf{M}}_C}}^{(i,j)})
\end{equation}
\begin{equation}\label{eq3}
	{\bf{F}}_{{{\bf{A}}_T}}^{(i,j)} = {{\bf{A}}_T}({\bf{F}}_{{{\bf{M}}_C}}^{(i,j)})
\end{equation}
\begin{equation}\label{eq4}
	{{\bf{F}}^{(i,j)}} = {\bf{Concat}}({\bf{F}}_{{{\bf{A}}_C}}^{(i,j)},{\bf{F}}_{{{\bf{A}}_T}}^{(i,j)})
\end{equation}

\subsection{The Feature Pyramid Fusion Head}
\begin{figure*}
	\centering
	\includegraphics[width=0.65\textwidth]{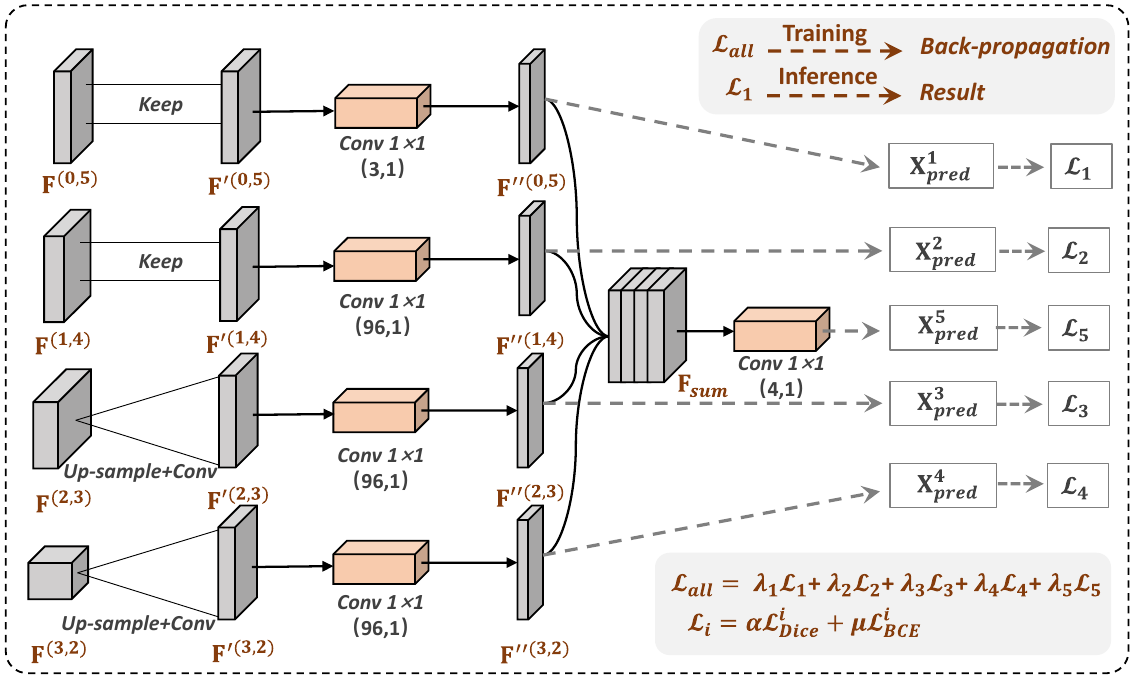}
	\caption{The structure of the feature pyramid fusion head. Where, we utilize four different scales of saliency maps as input for the head segment. Unlike traditional eight-neighborhood clustering segmentation, we directly employ a loss function to impose constraints and achieve end-to-end segmentation.}
	\label{fig4}
\end{figure*}

The overall structure of the Head part in IDNANet is shown in Fig. \ref{fig4}. For the selection of  ${{\bf{F}}^{(0,5)}}$, to retain the detailed features of the original channel, we choose the strategy of dimension preservation, so that the feature map will avoid the problem of information loss due to the lower adoption. We pool the deep information for the choice of ${{\bf{F}}^{(1,4)}}$ to complement the detailed features. ${{\bf{F}}^{(2,3)}}$ and ${{\bf{F}}^{(3,2)}}$ come from our mining of deeper features of the input image. Because the dimension changes during down-sampling, we need to pool the convolution operation on the channels and use up-sampling to recover the feature shape. In Fig. \ref{fig4}, ${{\bf{F'}}}$ is the feature map after the integration of different channels and dimensions. To obtain the single-channel information aggregation feature map, we perform a $Conv 1\times1$ channel merging operation ${{\bf{F''}}}$ on  to integrate the dimensions of all the channel incorporating feature maps  into ${96 \times H \times W}$. Meanwhile, we refer to ${{{\bf{F''}}}^{(0,5)}}$, ${{{\bf{F''}}}^{(1,4)}}$, ${{{\bf{F''}}}^{(2,3)}}$ and ${{{\bf{F''}}}^{(3,2)}}$ as ${\bf{X}}_{pred}^1\sim{\bf{X}}_{pred}^4$, respectively. Where ${\bf{X}}_{pred}^i$ represents the $i$-th predicted image and ${\bf{X}}_{pred}^i \in {{\mathbb{R}}^{1\times H \times W}}$. Considering that feature maps of different scales as results will have significant differences in training, we take ${\bf{X}}_{pred}^1\sim{\bf{X}}_{pred}^4$ as input, respectively, and obtain mixed feature map ${{\bf{F}}_{sum}}$ by channel incorporating. Finally, ${\bf{X}}_{pred}^5$ is output by channel integration. This Head design can ensure multi-scale feature information and reduce the impact of different scale targets on detection accuracy. In addition, we use ${\bf{X}}_{pred}^1\sim{\bf{X}}_{pred}^5$ in the training process and perform a weighted sum of the losses, allowing different loss value scales to adjust the network weights. But in the inference part, we only use the ${\bf{X}}_{pred}^1$.

\subsection{Weighted Dice Binary Cross-Entropy Loss}
In segment-based object detection networks, $IoU$ is a commonly used metric to measure the similarity between the predicted results and the true labels. However, such loss function also has some drawbacks for infrared small target detection. Therefore, we improve the loss function in DNANet called WD-BCE loss function. Dice loss can alleviate the negative impact of the imbalance between foreground and background in samples. The training of Dice loss pays more attention to the network mining the foreground region guaranteed to have a low False Negatives (FN). However, there will be loss saturation problem, and CE loss is to calculate the loss of each pixel equally. The loss function we designed is shown in Equation \ref{eq5}.

\begin{equation}\label{eq5}
	{{\cal L}_{all}} = {\lambda _1}{{\cal L}_1} + {\lambda _2}{{\cal L}_2} + {\lambda _3}{{\cal L}_3} + {\lambda _4}{{\cal L}_4} + {\lambda _5}{{\cal L}_5}
\end{equation}
Where, $\lambda _i$ is a trainable parameter initialized to a uniform distribution, ${\lambda _i} \sim U(0,1)$. Such parameters can be used to regulate different prediction output losses. $\lambda _i$ is the output loss of the $i$-th head, which is calculated as shown in Equation \ref{eq6}.
\begin{equation}\label{eq6}
	{{\cal L}_i} = \alpha {\cal L}_{Dice}^i + \mu {\cal L}_{BCE}^i
\end{equation}
Here, $\alpha$ and $\mu$ are the loss balance factors, which are used to adjust the dataset with unbalanced classes. When encountering a dataset with more categories, it can be adjusted appropriately. ${\cal L}_{Dice}^i$ and ${\cal L}_{BCE}^i$ are the Dice loss and BCE loss of the $i$-th output loss, respectively. 

\section{Proposed Dataset}
\subsection{Motivation}
To the best of our knowledge, most of the public datasets suffer from the following deficiencies. They lack an adequate amount of real data, resulting in limited coverage of target sizes. And the targets in existing synthetic datasets tend to be excessively realistic. The point annotations are usually too large, leading to strong contrast with the surrounding environment. While the background in these datasets is derived from real images, the targets are synthesized, often failing to achieve accurate thermal simulation. As a result, the thermal infrared images exhibit more uniform distribution across the surface. Previous datasets relied on manual labeling or ground truth (GT) synthesis, but these approaches have inherent limitations. Additionally, existing datasets are predominantly either entirely real or entirely synthetic. In contrast, our proposed dataset comprises both real targets and real data labels. And our dataset combines both real data with manual labels and a synthetic dataset with GT. Moreover, the target images are synthesized based on the collected real data. Consequently, training models on our dataset yields superior generalization. 

\subsection{The BIT-SIRST Dataset}
\begin{figure*}
	\centering
	\includegraphics[width=0.85\textwidth]{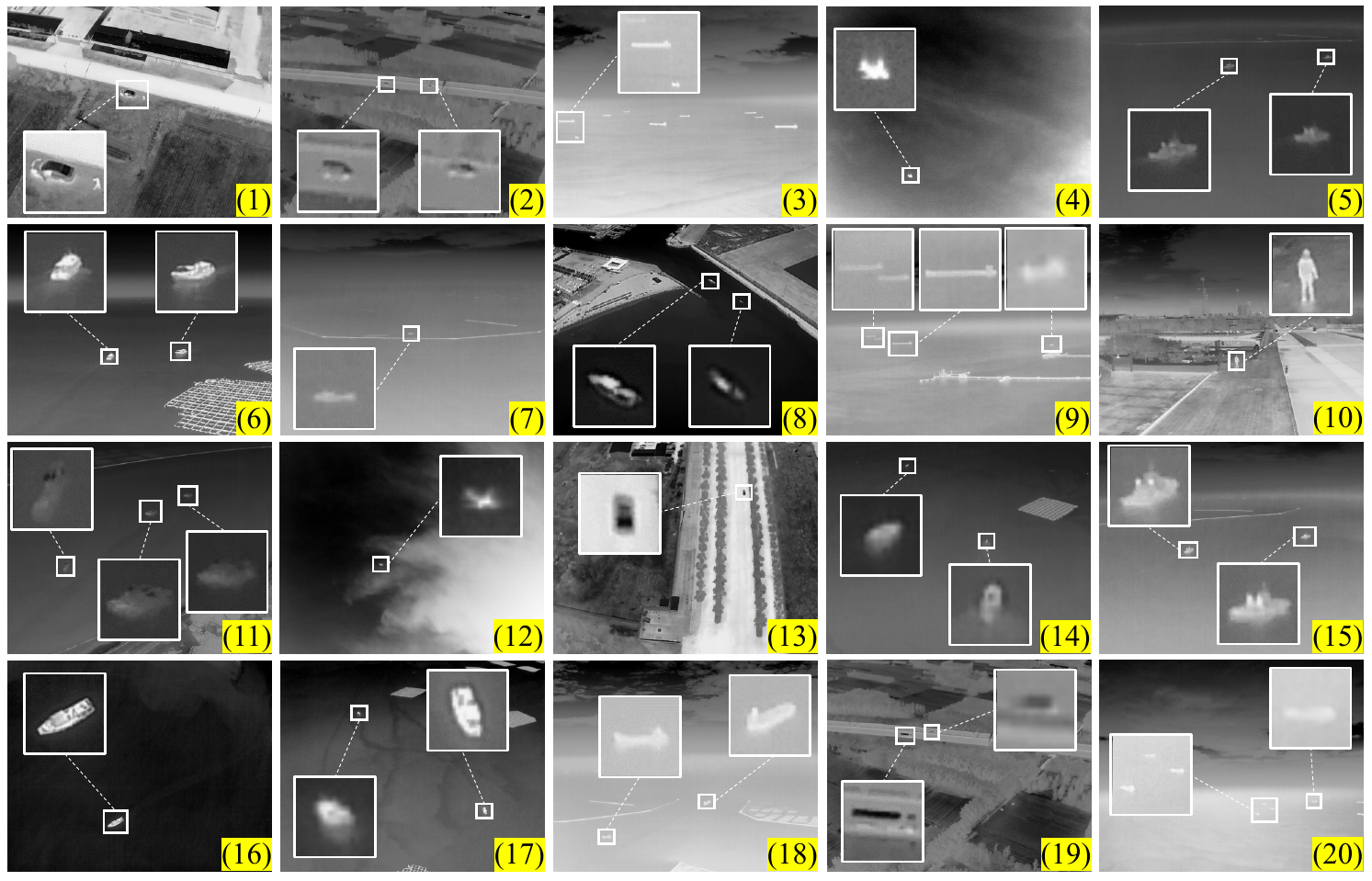}
	\caption{Representative infrared images from the BIT-SIRST dataset with various backgrounds. To enhance visibility, the demarcated area is enlarged, making it easier to see when zoomed in on a computer screen. The collected infrared small target images are numbered (1)-(20).}
	\label{fig5}
\end{figure*}

Unlike previous datasets for infrared small target detection that rely on local contrast, our dataset retains the contour information and has the added effect of local contrast. In the BIT-SIRST dataset, we select infrared small targets observed in daily life and special scenes like the sky, land, and sea. In the synthetic dataset, we gather data from real scenes. The real scene serves as the background, while the real target is adjusted using the infrared simulation method. This serves three main purposes. Firstly, it retains the information distribution and noise information from the real scene. Secondly, by collecting target information from the real scene, we can closely replicate the infrared characteristics of the target. Additionally, this method allows for manual intervention to manipulate the direction and pose of the target, thereby expanding the dataset. We combine synthetic data with the real collected data for training. This approach makes the model better suited for synthetic data and enhances its universality through training on real data. We assess the reliability of our dataset by conducting tests on the real collected data. In Table \ref{table1}, we compare our BIT-SIRST dataset with existing SIRST datasets. In Fig. \ref{fig5}, we select representative images from BIT-SIRST that contain various targets.

\begin{table*}[width=1\textwidth,cols=7,pos=h]
	\centering
	\caption{Main characteristics of several open SIRST datasets. Note that, our BIT-SIRST dataset contains common background scenes, various target types, and most ground truth annotations.}
	\label{table1}
	\begin{tabular}{c|c|c|c|c}
		\hline
		\textbf{Dataset} & \textbf{Image Type} & \textbf{Background Scene}       & \textbf{\#Image} & \textbf{Label Type}                \\ \hline
		SIRST(ACM)  & real                & Cloud/City/Sea                  & 427              & Manual Coarse Label                  \\ \hline
		NUST-SIRST       & synthetic           & Cloud/City/River/Road           & 10000            & Manual Coarse Label                  \\ \hline
		CQU-SIRST(IPI)   & synthetic           & Cloud/City/Sea                  & 1676             & Ground Truth                            \\ \hline
		NUDT-SIRST       & synthetic           & Cloud/City/Sea/Field/Highlight  & 1327             & Ground Truth                         \\ \hline
		\rowcolor[HTML]{D9D9D9} 
		BIT-SIRST(Ours)  & real/synthetic    & Cloud/City/Road/Field/River/Sea & 10568            & Manual Coarse Label/Ground Truth       \\ \hline
	\end{tabular}
\end{table*}

\section{Experiments}
\subsection{Evaluation Metrics}
\subsubsection{Intersection over Union ($IoU$)} 
$IoU$ is commonly used to evaluate the network's ability to describe shapes. Thus, we utilize $IoU$ to assess the network's performance, specifically the mean Intersection over Union ratio ($mIoU$). Equation \ref{eq7} provides the equation for calculating the $IoU$.

\begin{equation}\label{eq7}
	IoU = \frac{{{A_{inter}}}}{{{A_{union}}}}
\end{equation}
Where $A_{inter}$ and $A_{union}$ represent the interaction areas and all areas of prediction and label, respectively.
\subsubsection{Probability of Detection ($P_d$) and False-alarm Rate ($F_a$)} 
$P_d$ represents the network's ability to accurately detect the target, and $F_a$ represents the performance of generating false alarms during the detection process as shown in Equations \ref{eq8} and \ref{eq9}.
\begin{equation}\label{eq8}
	{P_d} = \frac{{{N_{correct}}}}{{{N_{all}}}}
\end{equation}
\begin{equation}\label{eq9}
	{F_a} = \frac{{{P_{false}}}}{{{P_{all}}}}
\end{equation}
Where $N_{correct}$ and $N_{all}$ represent the numbers of correctly predicted targets and all targets. $P_{false}$ and $P_{all}$ represent the numbers of falsely predicted pixels and all image pixels, respectively.
\subsubsection{Receiver Operation Characteristics (ROC)} 
The ROC curve takes the $F_a$ as the horizontal axis and the $P_d$ as the vertical axis. Generally, with the same false-alarm rate, the higher the detection rate, the better the algorithm's performance.

\subsection{Implementation Details}
Our experiments mainly use BIT-SIRST, NUDT-SIRST, SIRST and NUST-SIRST datasets. In the training and testing of IDNANet, we resize the input image to $256 \times 256$ resolution. In the comparison experiments with other SOTA models, we keep the configuration of the original model. The optimizer is Adagrad \citep{duchi2011adaptive} with a learning rate 0.05, and the batch size is set to 8 per GPU. All models are implemented in PyTorch \citep{paszke2019pytorch} on a computer with an Intel Core i9-10920X@3.50 GHz CPU and two Nvidia RTX3090 GPUs. Since IDNANet is designed to incorporate transformers, we extend the training epochs to 3000. For the parameters of WD-BCE, we set to $\alpha=1$ and $\mu=1$. The hyperparameter settings of the traditional method are shown in Table \ref{table2}.

\begin{table*}[t]
	\centering
	\caption{Detailed hyper-parameter settings of traditional methods for comparison.}
	\label{table2}
	\begin{tabular}{c|c}
		\hline
		\textbf{Methods}       & \textbf{Hyper-parameter settings}     \\ \hline
		Top-Hat               & Structure shape: disk, Size: 5$\times$5   \\
		Max-Median        & Window size = $\left\lbrace 3,5,7,9\right\rbrace $     \\
		RLCM                 & Window size = $\left\lbrace 3,5,7,9\right\rbrace $     \\
		WSLCM              & Cell size = $\left\lbrace 7, 9, 11\right\rbrace$    \\
		MPCM             & $N = 1,3, ... , 9$, threshold factor: $k = 13$     \\
		IPI                       & Patch size: 50$\times$50, stride: 20, $\lambda=L/min(m, n)^{1/2},L = 2.5, \varepsilon=10^{-7}$     \\
		NRAM               & Patch size: 30$\times$30, Slide step: 10, $\lambda = 1/\sqrt{min(m,n)}$     \\				
		\hline
	\end{tabular}
\end{table*}

\begin{table*}[width=1\textwidth,cols=7,pos=h]
	\centering
	\caption{$mIoU (\times 10^{-2})$, $P_d (\times 10^{-2})$ and $F_a (\times 10^{-6})$ values of different methods achieved on SIRST and NUDT-SIRST datasets.}
	\label{table3}
	\begin{tabular}{c|c|c|ccc|ccc}
		\hline
		\multicolumn{1}{c|}{}                                                & \multicolumn{1}{c|}{}                               & \multicolumn{1}{c|}{}                & \multicolumn{3}{c|}{\textbf{SIRST}}                                                                                                                                                                                                                     & \multicolumn{3}{c}{\textbf{NUDT-SIRST}}                                                                                                                                                                                                                     \\ \cline{4-9} 
		\multicolumn{1}{c|}{\multirow{-2}{*}{\textbf{Methods}}}              & \multicolumn{1}{c|}{\multirow{-2}{*}{\textbf{Description}}}      & \multicolumn{1}{c|}{\multirow{-2}{*}{\textbf{Year}}} & \multicolumn{1}{c|}{\textit{\textbf{$mIoU (\times 10^{-2})$}}}                                 & \multicolumn{1}{c|}{\textit{\textbf{$P_d (\times 10^{-2})$}}}                                   & \multicolumn{1}{c|}{\textit{\textbf{$F_a (\times 10^{-6})$}}}                                   & \multicolumn{1}{c|}{\textit{\textbf{$mIoU (\times 10^{-2})$}}}                                 & \multicolumn{1}{c|}{\textit{\textbf{$P_d (\times 10^{-2})$}}}                                   & \multicolumn{1}{c}{\textit{\textbf{$F_a (\times 10^{-6})$}}}                                   \\ \hline
		Top-Hat               & Filter-based       & 1996    & 7.14      & 79.84    & 1012.00      & 20.72     & 78.41      & 78.41     \\
		Max-Median       & Filter-based       & 1999    & 4.17      & 69.20    & 55.33          & 4.20       & 58.41      & 36.89     \\
		RLCM                  & HVS-based             & 2018     & 21.02    & 80.61    & 199.15        & 15.14     & 66.35      & 163.00    \\
		WSLCM              &HVS-based              & 2021     & 1.02      & 80.99    & 45846.16    & 0.85       & 74.60      & 52391.63  \\
		MPCM                & HVS-based             & 2016      & 12.38    & 83.27     & 17.77         & 5.86       & 55.87       & 115.96      \\
		IPI                        &IDS-based               & 2013      & 25.67    & 85.55    & 11.47          & 17.76     & 74.49       & 41.23        \\
		NRAM                & IDS-based              & 2018      & 12.16    & 74.52    & 13.85          & 6.93       & 56.4         & 19.27         \\
		MDvsFA            & DL-based               & 2019      & 61.77    & 92.40    & 64.90          & 45.38     & 86.03        & 200.71     \\
		ISNet                   & DL-based               & 2022      & 72.04    & 94.68     & 42.46         & 71.27     & 96.93        & 96.84       \\
		ACM                   & DL-based               & 2021      & 64.92    & 90.87     & 12.76         & 57.42     & 91.75        & 39.73       \\
		ALCNet              & DL-based               & 2021      & 67.91    & 92.78     & 37.04         & 61.78     & 91.32        & 36.36        \\
		DNANet             & DL-based               & 2022      & 76.86    & 96.96     & 22.50         & 87.42     & 98.31        & 24.50        \\
		UIU-Net              & DL-based               & 2023      & 69.90    & 95.82     & 51.20         & 75.91     & 96.83        & 18.61       \\ 
		\rowcolor[HTML]{D9D9D9} 
		\cellcolor[HTML]{D9D9D9}\textbf{IDNANet (Ours)} & \cellcolor[HTML]{D9D9D9}\textbf{DL-based} & \cellcolor[HTML]{D9D9D9}\textbf{2023} & \cellcolor[HTML]{D9D9D9}{\color[HTML]{B22222} \textbf{79.72}} & \cellcolor[HTML]{D9D9D9}{\color[HTML]{B22222} \textbf{98.52}} & \cellcolor[HTML]{D9D9D9}{\color[HTML]{B22222} \textbf{2.32}} & \cellcolor[HTML]{D9D9D9}{\color[HTML]{B22222} \textbf{90.89}} & \cellcolor[HTML]{D9D9D9}{\color[HTML]{B22222} \textbf{98.23}} & \cellcolor[HTML]{D9D9D9}{\color[HTML]{B22222} \textbf{3.24}} \\  \hline
	\end{tabular}
\end{table*}

\begin{figure*}
	\centering
	\includegraphics[width=1\textwidth]{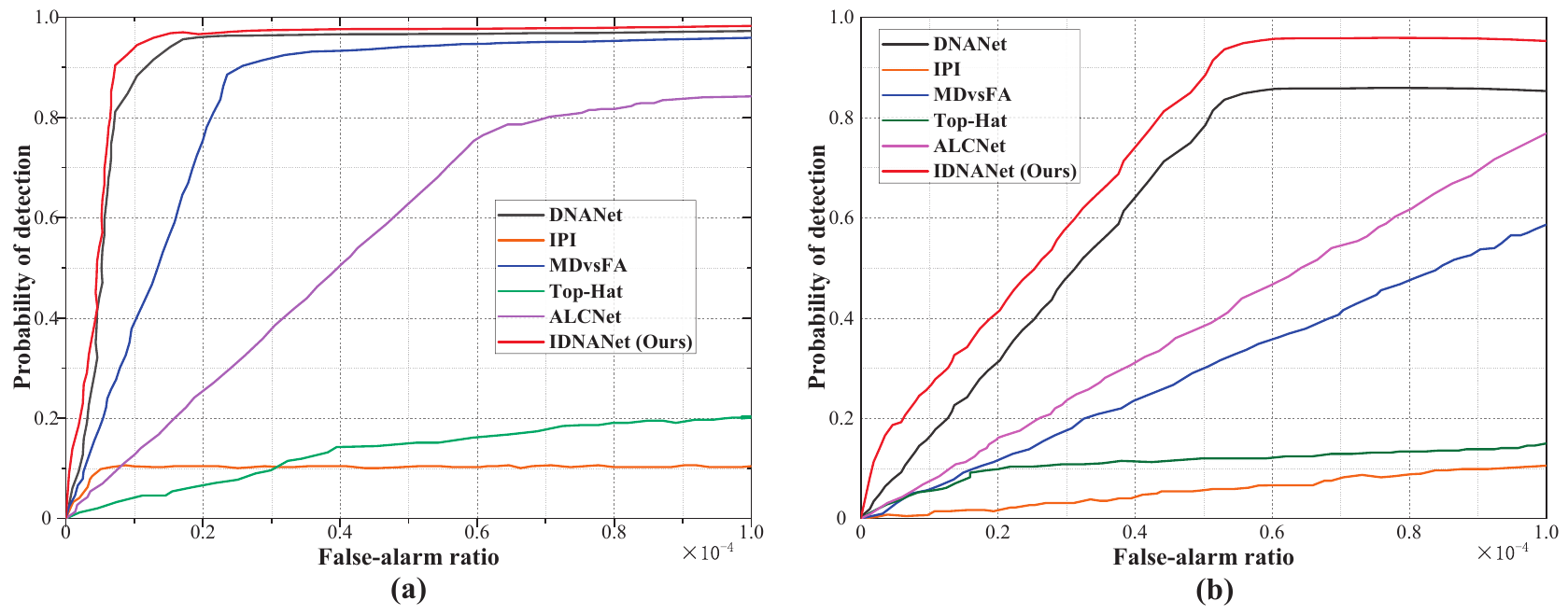}
	\caption{ROC curves of different methods on the (a) NUDT-SIRST dataset and (b) SIRST dataset. Our IDNANet performance remains stable on both dataset NUDT-SIRST and SIRST. Additionally, the ROC curve of IDNANet is located in the top left corner.}
	\label{fig6}
\end{figure*}

\subsection{Comparison to the State-of-the-art Methods}
\subsubsection{Performance on open datasets}
\begin{figure*}
	\centering
	\includegraphics[width=1\textwidth]{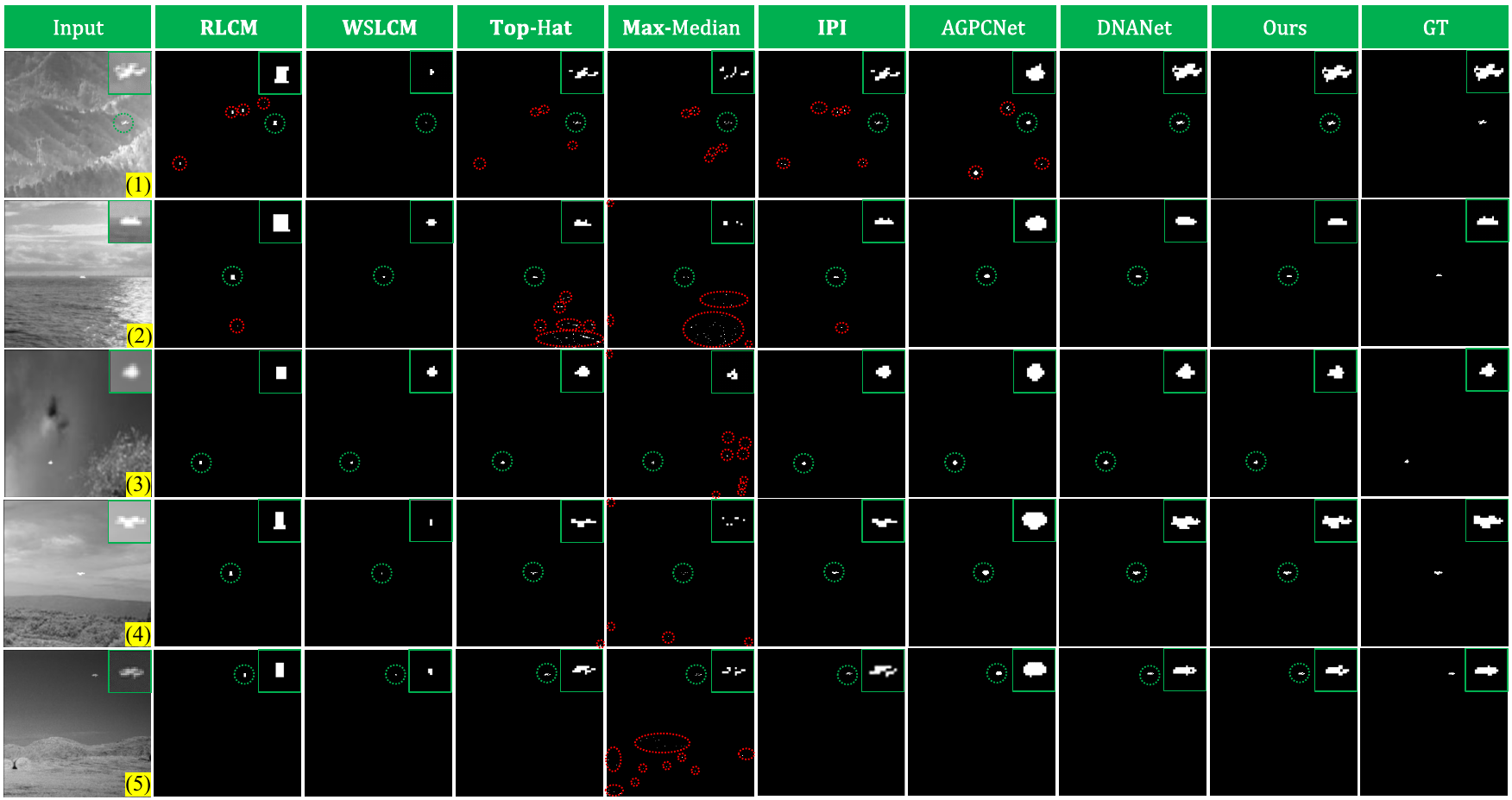} 
	\caption{Visual results achieved by RLCM, WSLCM, Top-Hat, Max-Median, IPI, AGPCNet, DNANet and our network for different infrared small targets. The correctly detected target, false alarm detection areas are highlighted by green and red dotted circle.}
	\label{fig7}
\end{figure*}

\begin{figure*}
	\centering
	\includegraphics[width=1\linewidth]{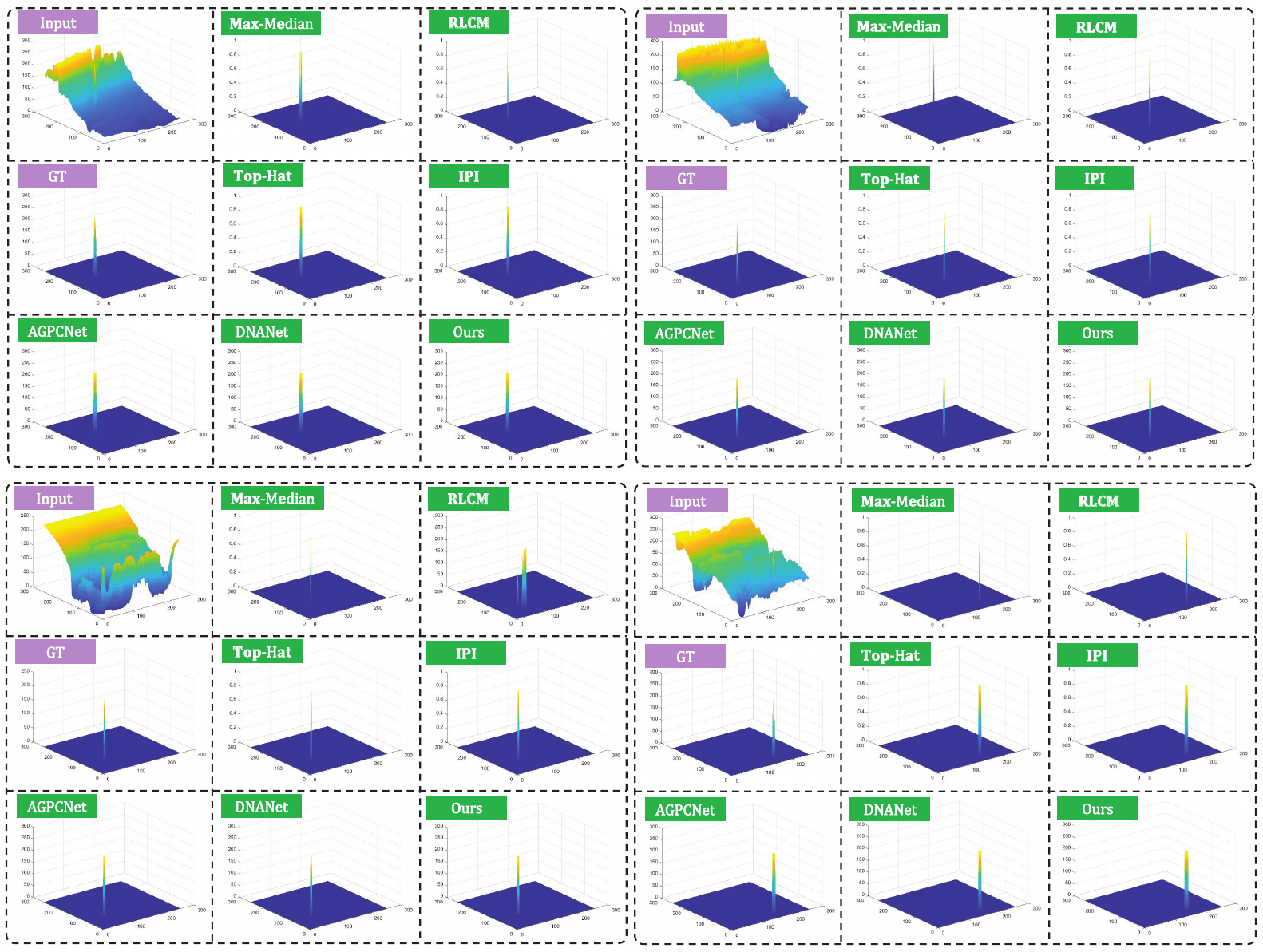}
	\caption{3D visualization results of different methods in the BIT-SIRST dataset.}
	\label{fig8}
\end{figure*}

To verify the superiority of the proposed method, we contrast some SOTA models. Traditional methods include Top-Hat, Max-Median, RLCM \citep{han2018infrared}, WSLCM, MPCM \citep{wei2016multiscale}, IPI and NRAM \citep{zhang2018infrared}, and the deep learning based methods MDvsFA, ISNet \citep{zhang2022isnet}, ACM, ALCNet, DNANet and UIU-Net on the SIRST and NUDT-SIRST. We keep all remaining parameters the same as their original papers. In terms of quantitative experiments, the results of our comparative experiments are shown in Table \ref{table3}. From Table \ref{table3}, we can also see clearly that IDNANet significantly outperforms the traditional method in performance. Because the conventional methods rely on the manually set threshold, such a setting has substantial limitations for infrared small target detection. The deep learning-based method outperforms the traditional methods in $mIoU$ and the conventional methods in detection and false alarm rate performance. This also verifies that the data-driven method is better than the model-driven detection algorithm. Compared with these deep learning-based methods, the improvement made by our proposed IDNANet method is also apparent. On the SIRST dataset, our IDNANet can achieve 79.72\% on $mIoU$, 98.52\% for $P_d$, and 2.32\% for $F_a$. Compared with our improved base network DNANet, our $mIoU$ is improved by 2.86\%. On the NUDT-SIRST dataset, our IDNANet can reach 90.89\% on $mIoU$, 98.23\% on $P_d$, and 3.24\% on $F_a$. Compared with DNANet, our $mIoU$ is improved by 3.47\%  on the NUDT-SIRST dataset. This is because we add the Swin-T feature extraction network to DNANet to effectively capture long-distance dependencies and global features. It is also because our ACmix block enhances the infrared small target features. Quantitative experiments demonstrate the effectiveness of the improvements of IDNANet. Fig. \ref{fig6} shows the ROC curves we compared in our experiments. We can observe that the ROC curve ($F_a$, $P_d$) of IDNANet is closer to the upper left corner than other methods. Therefore, our IDNANet outperforms the other methods in performance. At the same time, it also proves that our IDNANet has a better effect in infrared small target detection and background suppression. Fig. \ref{fig7} shows visual results achieved by different methods for infrared small targets. Fig. \ref{fig8} present qualitative 3-D visualization results of different detection methods on BIT-SIRST datasets.

\subsubsection{Performance on the BIT-SIRST datasets} 
To validate the effectiveness of the proposed BIT-SIRST dataset, we train and test the popular deep learning methods using BIT-SIRST as shown in Table \ref{table4}. Based on Table \ref{table4}, it is evident that our IDNANet can achieve state-of-the-art performance comparable to other models. Specifically, the $mIoU$ of IDNANet can reach 80.76\%, the $P_d$ can reach 98.64\%, and the $F_a$ can reach 19.24\% at maximum.

\begin{table*}[t]
	\centering
	\caption{$mIoU ( \times 10^{-2})$, $P_d ( \times 10^{-2})$ and $F_a ( \times 10^{-6})$ values of IDNANet achieved on BIT-SIRST dataset.}
	\label{table4}
	\begin{tabular}{c|c|c|ccc}
		\hline
		\textbf{Methods}       & \textbf{Description} & \textbf{Year} & $mIoU ( \times 10^{-2})$   & $P_d ( \times 10^{-2})$    & $F_a ( \times 10^{-6})$    \\ \hline
		AGPCNet       & DL-based  & 2021 & 67.11 & 89.78 & 37.04 \\
		DNANet        & DL-based   & 2022 & 76.07 & 96.96 & 22.5  \\
		MTU-Net       & DL-based   & 2023 &  77.32     & 97.25      &  20.16     \\
		\rowcolor[HTML]{D9D9D9} IDNANet(Ours) & DL-based   & 2023 & \cellcolor[HTML]{D9D9D9}{\color[HTML]{B22222} \bf{80.76}} & \cellcolor[HTML]{D9D9D9}{\color[HTML]{B22222} \bf{98.64 }}     &  \cellcolor[HTML]{D9D9D9}{\color[HTML]{B22222} \bf{12.24}}     \\ \hline
	\end{tabular}
\end{table*}

\subsection{Abliation Study}
To verify the effectiveness of IDNANet, we conduct a series of ablation experiments. Through the analysis of experimental results, we demonstrate the influence of ACmix block, WD-BCE, the number of output results and Swin-T backbone on the performance of IDNANet.

\subsubsection{Efficacy of ACmix Block}
In this section, we perform an experimental study on the ACmix block added to IDNANet. The experiments aim to confirm that the inclusion of $\bf{AB}$ has a significant impact on improving the network's $mIoU$. In IDNANet, we train separately on the NUDT-SIRST dataset by adding different numbers of $\bf{AB}$. It is important to note that we replace it with $\bf{RCB}$ in the sections where $\bf{AB}$ is not added. The experimental results are presented in Table \ref{table5}. Table \ref{table5} clearly shows that the network's accuracy is lowest when IDNANet does not include the $\bf{AB}$ module. As the number of $\bf{AB}$ modules increases, the $mIoU$ also gradually improves. When all four $\bf{AB}$ modules are added, the $mIoU$ reaches 90.89\%.

\begin{table*}[t]
	\centering
	\caption{Impact of varying numbers of ACmix blocks (AB) on the performance of IDNANet. In particular, the dataset is the NUDT-SIRST.}
	\label{table5}
	\begin{tabular}{c|c|c|c|c}
		\hline
		$\bf{AB1}$&  $\bf{AB2}$  &$\bf{AB3}$  &$\bf{AB4}$  & $mIoU ( \times 10^{-2})$                          \\ \hline
		\ding{55}&  \ding{55}&  \ding{55}&  \ding{55}& 84.60                         \\
		\ding{51}&  &  &  & 83.29                         \\
		\ding{51}&  \ding{51}&  &  & 82.02                         \\
		\ding{51}&  \ding{51}&  \ding{51}&  & 88.04                         \\
		\cellcolor[HTML]{D9D9D9}\ding{51}&  \cellcolor[HTML]{D9D9D9}\ding{51}&  \cellcolor[HTML]{D9D9D9}\ding{51}& \cellcolor[HTML]{D9D9D9}\ding{51}& \cellcolor[HTML]{D9D9D9}{\color[HTML]{B22222} \bf{90.89}} \\ \hline
	\end{tabular}
\end{table*}

\subsubsection{Efficacy of loss function} 
We assess the impact of the WD-BCE loss function on different methods, and the experimental results are presented in Table \ref{table6}. We compare the performance of DNANet, AGPCNet, and MTU-Net when using alternative loss functions. It is important to note that the configurations used in the comparison experiments are the original default settings of the network. To ensure optimal network convergence, we set the epoch to 1500 for all experiments. When the network outputs only one result, such as MTU-Net for loss calculation, $\lambda_i$ in our WD-BCE is set to 1. From Table \ref{table6}, we can see that relative to the original loss function, in DNANet, our designed WD-BCE improves the $mIoU$ by 1.03\%. In AGPCNet, $mIoU$ is improved by 1.80\%. In MTU-Net, $mIoU$ is boosted by at most 1.68\%. This proves that WD-BCE plays an essential role in improving performance.

\begin{table*}
	\centering
	\caption{Experimental results after replacing the loss functions of different networks with WD-BCE.}
	\label{table6}
	\begin{tabular}{c|c|c|c}
		\hline
		\textbf{Methods}         & \textbf{Loss Function} & \textbf{Dataset}                & \textbf{$mIoU (\times 10^{-2})$} \\ \hline
		\multirow{2}{*}{DNANet}  & SoftIoU                & \multirow{2}{*}{NUST-SIRST}     & 86.60              \\ \cline{2-2} \cline{4-4} 
		& WD-BCE                 &                                 & \color[HTML]{B22222} \textbf{87.63}     \\ \hline
		\multirow{2}{*}{AGPCNet} & SoftIoU                & \multirow{2}{*}{NUDT-SIRST}     & 44.64              \\ \cline{2-2} \cline{4-4} 
		& WD-BCE                 &                                 & \color[HTML]{B22222} \textbf{46.44}     \\ \hline
		\multirow{2}{*}{MTU-Net} & FocalIoU               & \multirow{2}{*}{NUDT-SIRST-Sea} & 60.55              \\ \cline{2-2} \cline{4-4} 
		& WD-BCE                 &                                 & \color[HTML]{B22222} \textbf{62.23}     \\ \hline
	\end{tabular}
\end{table*}

\begin{table}[t]
	\centering
	\caption{Test results of adding different losses on IDNANet. In particular, the dataset is the NUDT-SIRST.}
	\label{table7}
	\begin{tabular}{c|c|c|c|c|c}
		\hline
		${\cal L}os{s_1}$&  ${\cal L}os{s_2}$ &${\cal L}os{s_3}$ &${\cal L}os{s_4}$  & ${\cal L}os{s_5}$  & $mIoU (\times 10^{-2})$                           \\ \hline
		\cellcolor[HTML]{D9D9D9}\ding{51}&  \cellcolor[HTML]{D9D9D9}\ding{51}& \cellcolor[HTML]{D9D9D9} \ding{51}&  \cellcolor[HTML]{D9D9D9}\ding{51}& \cellcolor[HTML]{D9D9D9}\ding{51} & \cellcolor[HTML]{D9D9D9}{\color[HTML]{B22222} \bf{90.89}}                        \\
		&  \ding{51}&  \ding{51}&  \ding{51}& \ding{51} & 88.88                      \\
		&                &  \ding{51}&  \ding{51}& \ding{51}  & 88.49                      \\
		&                &                &  \ding{51}& \ding{51}  & 88.27                      \\
		&                &                &                & \ding{51}  & 87.88                      \\ \hline	               
	\end{tabular}
\end{table}

\subsubsection{Loss function selection of different weights}
In the paper, we provide five loss values as the default for the final loss function. During training, we back-propagate the error by taking a weighted sum of these five loss values. However, a question arises: Do all five losses require error back-propagation, or is there a distinction in terms of weighting? So we experiment with different numbers of losses. The experimental results are shown in Table \ref{table7}. ${\cal L}os{s_i}$ is calculated as shown in Equation \ref{eq10}.
\begin{equation}\label{eq10}
	\begin{array}{*{20}{c}}
		{{\cal L}os{s_i} = {\lambda _i}{{\cal L}_i},}&{i = 1,2,3,4,5}
	\end{array}
\end{equation}
From Table \ref{table7}, it is evident that the lowest $mIoU$ is 87.88\% when only the ${\bf{X}}_{pred}^5$ is used for calculating the loss value during loss backpropagation training. However, as more preds are used and more losses and components are computed, the $mIoU$ tends to improve. The highest achieved $mIoU$ is obtained when all the losses are used for calculating the total loss value. This results in a 3.01\% improvement in $mIoU$ compared to using only one ${\bf{X}}_{pred}$. Therefore, in our IDNANet, all losses are included in the calculation of the loss value.

\subsubsection{Performance comparison of different backbones}
To verify the effectiveness of the transformer backbone, we conduct experiments by replacing the IDNANet backbone with ResNet-34 and VGG10. The performance on the NUDT-SIRST dataset is shown in Table \ref{table8}. From Table \ref{table8}, we can also observe that IDNANet exhibits good $mIoU$ performance under the feature extraction of transformers. This further demonstrates that a transformer-based backbone is more beneficial for extracting features from small infrared targets.

\begin{table}[t]
	\centering
	\caption{$mIoU ( \times 10^{-2})$ values of IDNANet with different backbones on NUDT-SIRST dataset. In this study, we conduct a comparative analysis of the Swin-T v2 with prominent backbone networks including VGG10 and ResNet32. These architectures are commonly employed in infrared small target detection networks.}
	\label{table8}
	\begin{tabular}{c|c|c|c}
		\hline
		\textbf{Methods}       & \textbf{Backbone} & \textbf{Dataset} & $mIoU ( \times 10^{-2})$     \\ \hline
		IDNANet       & Swin-T v2  & NUDT-SIRST & 90.89  \\
		IDNANet        & ResNet-34   & NUDT-SIRST & 87.07   \\
		IDNANet       & VGG10   & NUDT-SIRST &  85.34    \\ \hline
	\end{tabular}
\end{table}

\section{Conclusion}
In this paper, we propose an improved IDNANet based on transformer. Different from existing deep learning-based SIRST detection methods, we introduce transformer to enhance the long-distance feature extraction capability on the basis of DNANet. Additionally, we enhance the multi-level feature interaction capability by designing the structure of ACmix block. We design the WD-BCE loss function to alleviate the negative impact caused by foreground-background imbalance. Moreover, we develop an open BIT-SIRST dataset. Experimental results on both public datasets and our dataset demonstrate that our proposed approach outperforms the state-of-the-art methods.

\bibliographystyle{elsarticle-harv}


\bibliography{IDNANet}


\end{document}